\documentclass[graybox]{svmult}

\usepackage[super]{nth}

\usepackage{type1cm}        
\usepackage{makeidx}         
\usepackage{graphicx}        
\usepackage{multicol}        
\usepackage[bottom]{footmisc}

\usepackage{newtxtext}       %
\usepackage{newtxmath}       

\makeindex             

\begin{document}

\title*{Predicting User-specific Future Activities using LSTM-based Multi-label Classification}
\author{Mohammad Sabik Irbaz, Fardin Ahsan Sakib and Lutfun Nahar Lota}
\institute{Mohammad Sabik Irbaz \at Islamic University of Technology, Gazipur, Bangladesh, \email{sabikirbaz@iut-dhaka.edu} \\ \\
Fardin Ahsan Sakib \at George Mason University, Fairfax, Virginia, USA, \email{fsakib@gmu.edu} \\ \\ 
Lutfun Nahar Lota \at Islamic University of Technology, Gazipur, Bangladesh, \email{lota.nahar.407@gmail.com}}
%
%
\maketitle

\abstract{User-specific future activity prediction in the healthcare domain based on previous activities can drastically improve the services provided by the nurses. It is challenging because, unlike other domains, activities in healthcare involve both nurses and patients, and they also vary from hour to hour. In this paper, we employ various data processing techniques to organize and modify the data structure and an LSTM-based multi-label classifier for a novel 2-stage training approach (user-agnostic pre-training and user-specific fine-tuning). Our experiment achieves a validation accuracy of 31.58\%, precision 57.94\%, recall 68.31\%, and F1 score 60.38\%. We concluded that proper data pre-processing and a 2-stage training process resulted in better performance. This experiment is a part of the "Fourth Nurse Care Activity Recognition Challenge" by our team "Not A Fan of Local Minima".}

\section{Introduction}

Activity prediction from sensor data is a relatively untapped field in machine learning. It has many real-life applications in domains like healthcare, transportation, and so on. This topic is mainly studied in Computer Vision and activities were recognized by visual data from the video feed. But it is computationally expensive and not feasible to implement on large scale. In ubiquitous computing, however, if we can collect the data from commonly available sensors like accelerometers and use the sensor data from activity recognition, which makes it is eligible for using numerous fields. The first three nurse care activity challenge dealt with human activity recognition from sensor data from the real world. The fourth challenge takes it one step ahead and deals with the next activity prediction from sensor data, which will enable caregivers to provide services efficiently to the patients. 

In the healthcare domain, due to the lack of datasets and complex interpretability and explainability of the sensor data, nurse care activity prediction had always been a challenging field. These datasets generally comes up with some common challenges: a lot of noisy and unreliable samples, overlapping time frame, irregular sampling rate, inconsistency between lab and field data and so on \cite{inoue2019nurse}. That's why, processing these datasets are often very tricky. Furthermore, predicting the future activities of a particular user from the previous activities is even more challenging.

To address these problems, the fourth \cite{vchd-s336-22} nurse care activity recognition challenge aims to deal with nurse or caregiver activity prediction from previous activities of a particular user. The goal of this challenge is to make the task scheduling of the nurses or caregivers easier by predicting the probable activities of the next hour based on their activities in previous hours. The challenge provides raw data which contain the records of nurse care activity in a particular hour with a start and end time.

We pre-process the raw data so that it can be used for the multi-label next activity prediction task. We propose a model architecture that leverages LSTM and Bi-LSTM as the backbone. Using our novel 2-stage training approach, i.e., user agnostic pre-training and user-specific fine-tuning, we achieve the best performance.

\section{Related Works}

Activity Recognition on health care data has not been studied much due to the complexity and limited availability of the dataset. And the works are mostly focused on patient activities \cite{espinilla2018ucami, lago2017contextact, stikic2011weakly}. This task is unique because it not only depends on the nurse that performs the activity but also on the patient who receives it. The first nurse care activity challenge collected a multi-modal dataset in a controlled environment \cite{inoue2019nurse}. Kadir et al. \cite{10.1145/3460418.3479391} used kNN classifier and achieved  87\% accuracy on 10-fold cross-validation and 66\% accuracy on leave-one-subject-out cross-validation. Haque et al. \cite{10.1145/3341162.3344848} used a GRU-based approach with an attention mechanism and achieved 66.43\% validation accuracy for person-wise one leaving out cross-validation. Although in the test data, the accuracy was lower. 

The second nurse care activity recognition challenge \cite{10.1145/3410530.3414611} used both lab and real-world data for activity recognition. This partially solved the weakness of the first challenge as it reduces the gap between lab simulation and the real world. Rasul et. al \cite{10.1145/3410530.3414335} used a basic imputation strategy for preprocessing, used a CNN model, and achieved a validation accuracy of 91.59\%. Although in the test data, the accuracy was lower. Irbaz et. al \cite{irbaz2020nurse} achieved a validation accuracy of 75\% and test accuracy of 22.35\%. They used both a high pass and a low pass filter to shape the data in the spatial domain during preprocessing and used the kNN classifier instead of a deep learning algorithm. They concluded that traditional machine learning techniques can also be quite useful in activity recognition. 

The third care nurse care activity recognition challenge \cite{10.1145/3351244} focused on recognizing the activities of a nurse based on the accelerometer data as it is the cheapest and the most feasible way of collecting activity data. Using random forest, Sayem et al \cite{10.1145/3460418.3479388} achieved a validation accuracy of 72\%. But in the test data, the highest accuracy achieved was 12.97\%. Although when test data became available, accuracy rose up to as much as 80\% for a few activities. 


\section{Datasets}
Dataset from the \nth{3} \cite{hj46-zs46-21} and \nth{4} \cite{vchd-s336-22} Nurse Care Activity Recognition Challenge are used for this research. As both of the datasets are part of a same dataset, the structure of the datasets are almost similar. Data collection process for the datasets are same \cite{inoue2019integrating}. The datasets contain total 28 labels or activity types performed by caregivers or nurses. 

\subsection{Third Nurse Care Activity Recognition Challenge Dataset}
\begin{figure}[h]
    \centering
    \includegraphics[width=11 cm]{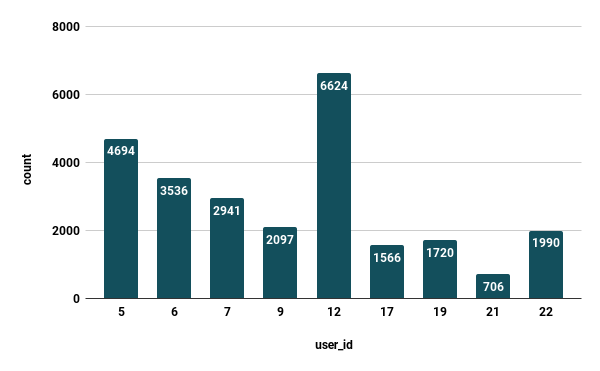}
    \caption{Third Nurse Care Data Distribution}
    \label{3rd_dist}
\end{figure}

There are 3 tabular datasets. Since our challenge task only focused on the next activity type prediction, we could only utilize the activity data \textit{(label\_train.csv)}. The size of the dataset is 27,448. In this dataset, they have 3 types of users: nurses, managers, and care managers. We only took the nurse user data which was more relevant to the \nth{4} Nurse Care challenge objective. The size of the nurse user data is 25,874. There are 9 nurse users in total: 5, 6, 7, 9, 12, 17, 19, 21, 22. In Figure \ref{3rd_dist}, we can see the distribution of this data which is very unbalanced.

\subsection{Fourth Nurse Care Activity Recognition Challenge Datasets}
\begin{figure}[h]
    \centering
    \includegraphics[width=10 cm]{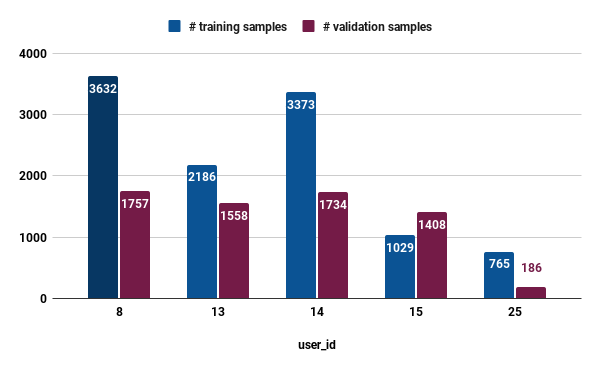}
    \caption{Fourth Nurse Care Data Distribution}
    \label{4th_dist}
\end{figure}

There are 5 different sets of data. We are using \textit{Care Record Data.zip} as training set and \textit{Test Data.zip} as validation set. Both of them have 5 users: 8, 13, 14, 15, and 25. The training set contains 10,985 samples and the validation set contains 6,643 samples. In Figure \ref{4th_dist}, the distribution of both the training and validation datasets are shown with respect to the users. We can see that user 25 has the lowest number of both types of samples.

\subsection{Challenges in Nurse Care Datasets}
We faced several challenges when working with the datasets. Two major challenges were pre-processing the raw datasets and working with imbalanced and noisy data.

The raw datasets of both challenges only have user\_id, date, hour, and activity type per record. Even though the goal of the \nth{4} Nurse Care challenge is to predict the activities of an hour based on the activities of previous hours, the datasets are not organized for that particular goal. So, they required some manipulation and pre-processing to be used as a sequence prediction dataset.  

From Figure \ref{3rd_dist} and \ref{4th_dist}, we can conclude that the datasets are imbalanced. It contains 6,624 training samples for user 12, but only 706 samples for user 21. Moreover, we also found out that the training and validation datasets were inconsistent with each other which results in lower accuracy and F1 score than expected. 

\section{Proposed Methodology}
Our proposed methodology can be divided into three steps. They are data pre-processing, user-agnostic pre-training, and user-specific fine-tuning. In the case of the user-agnostic pre-training approach, we do not specify or give preference to any users. But, we only provide specific user data for user-specific fine-tuning.

\subsection{Data Preprocessing}
We pre-processed and organized the raw datasets to be used for the multi-label next activity prediction task. All user data from \nth{3} and \nth{4} Nurse Care challenges have been pre-processed in a similar manner. First of all, for each user and a particular date and hour, all the activities are listed sequentially based on the starting time of those activities. For a specific date and hour, the unique activities are also listed. Secondly, if there is only one instance of a particular date, we remove that record from the dataset because it does not help us in predicting the activities of the next hour. To build the final dataset, we gather the hourly tasks of a particular date and iterate through them. During the iteration, the pointer indicates the unique tasks for the next hour, and the task in the preceding hours is considered the previous task. Table \ref{eg_proc} provides some examples of the data after pre-processing.

\begin{table}[h]
\centering
\begin{tabular}{c|c|c|c|c|c|c}
\textbf{user\_id} & \textbf{\begin{tabular}[c]{@{}c@{}}year-month\\ -date\end{tabular}} & \textbf{\begin{tabular}[c]{@{}c@{}}previous\\ hours\end{tabular}} & \textbf{\begin{tabular}[c]{@{}c@{}}previous\\ activities\end{tabular}} & \textbf{\begin{tabular}[c]{@{}c@{}}previous\\ activities\\ unique\end{tabular}} & \textbf{next\_hour} & \textbf{\begin{tabular}[c]{@{}c@{}}next\\ activities\end{tabular}} \\ \hline
\textbf{25} & 2018-05-30 & {[}7, 8{]} & \begin{tabular}[c]{@{}c@{}}{[}{[}10, 23, 6, 6, 6, 6{]}, \\ {[}6{]}{]}\end{tabular} & {[}{[}6, 10, 23{]}, {[}6{]}{]} & 9 & {[}10{]} \\ \hline
\textbf{25} & 2018-05-26 & {[}6, 7, 8{]} & \begin{tabular}[c]{@{}c@{}}{[}{[}23{]}, {[}6, 10, \\ 10, 6, 6, 6{]}, {[}18, 6{]}{]}\end{tabular} & \begin{tabular}[c]{@{}c@{}}{[}{[}23{]}, {[}6, 10{]}, \\ {[}6, 18{]}{]}\end{tabular} & 9 & {[}7, 18, 22{]} \\ \hline
\textbf{25} & 2018-06-03 & \begin{tabular}[c]{@{}c@{}}{[}7, 8, \\ 10, 11{]}\end{tabular} & \begin{tabular}[c]{@{}c@{}}{[}{[}6, 6, 6, 23, 10, 6{]}, {[}6{]}, \\ {[}10, 24{]}, {[}6, 6, 6{]}{]}\end{tabular} & \begin{tabular}[c]{@{}c@{}}{[}{[}6, 10, 23{]}, {[}6{]}, \\ {[}10, 24{]}, {[}6{]}{]}\end{tabular} & 12 & {[}6{]} \\ \hline
\textbf{25} & 2018-06-03 & {[}7{]} & {[}{[}6, 6, 6, 23, 10, 6{]}{]} & {[}{[}6, 10, 23{]}{]} & 8 & {[}6{]} \\ \hline
\textbf{25} & 2018-06-03 & {[}7, 8{]} & \begin{tabular}[c]{@{}c@{}}{[}{[}6, 6, 6, 23, \\ 10, 6{]}, {[}6{]}{]}\end{tabular} & {[}{[}6, 10, 23{]}, {[}6{]}{]} & 10 & {[}10, 24{]} \\ \hline
\end{tabular}
\caption{Examples of Pre-processed Data}
\label{eg_proc}
\end{table}

\begin{figure}[h]
    \centering
    \includegraphics[width=5.5 cm]{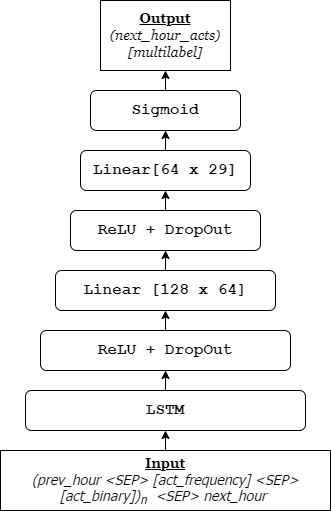}
    \caption{Model architecture for pre-training and fine-tuning}
    \label{arch}
\end{figure}

\subsection{User-agnostic Pre-training}
We used LSTM \cite{hochreiter1997long} and BiLSTM \cite{schuster1997bidirectional} as backbone of our model architecture to train the model. Both \nth{3} and \nth{4} Nurse Care Challenge data for this pre-training without specifying or giving preference to any user. The processed data is reconstructed in a particular format to utilize the sequential long-term and short-term dependencies of LSTM. For each previous hour in the processed data shown in Table \ref{eg_proc}, we took the hour, added a separator, an array of activity type frequencies of that hour, another separator, and, finally, another array of activity type binary which indicates if an activity among the 28 was conducted in that hour. After adding all of the hourly activities in a sequence, we added another separator and concatenated the next hour with it. This input is passed to an LSTM layer. The output of the LSTM is passed to a ReLU and a dropout layer and again passed into a Linear Fully connected layer. The result of the Linear layer goes through the same process again, and finally, the logit values are passed to a sigmoid layer to generate the output predictions. We get an array as output that specified if a particular type of activity was conducted or not. We also conducted another experiment replacing LSTM with BiLSTM. Figure \ref{arch} provides a detailed architecture of the modeling approach.

\subsection{User-specific Fine-tuning}
The pre-trained model is fine-tuned on the \nth{4} Nurse Care Challenge Data. We conduct user-specific fine-tuning for all 5 users. All the input processing and modeling approach are kept the same.

\section{Experimental Analysis}
To train and evaluate our proposed methodology, we did a couple of experiments in 2 stages. In the first stage, we trained an LSTM and BiLSTM-based model on 4th Nurse Care Data and later on both 3rd and 4th Nurse care data without specifying users. In the second stage, we used the pre-trained models of the first stage and fine-tuned them using specific user data (user 8, 13, 14, 15, and 25) separately. 

To evaluate the performances of each stage and each modeling approach, we considered accuracy (exact match), precision, recall, and F1 score. The accuracy or exact match is the strictest metric for multi-label classification because the result drastically decrease even if there is one mismatch of a particular record during inference. F1-score provides a more balanced result considering the precision and recall and giving some reward for some matches. That's why, in most multi-label classification tasks, F1 score is much higher than the accuracy.

\subsection{Experiment Setup}
We trained all our models using a RAM of 12.68 GB, Intel(R) Xeon(R) CPU @ 2.30GHz, and Tesla P100-PCIE GPU. The batch size was set to 2, and the maximum sequence length to 1200. We set the dropout rate to 0.1 and the learning rate to 4e-4. The models were trained for 50 epochs. AdamW \cite{kingma2014adam, loshchilov2017decoupled} was used as the optimizer, and Binary Cross Entropy with logit loss as the loss function. The threshold was set to 0.5 which means if the sigmoid value is greater than or equal to 0.5 after inference, we reset it to 1 or else 0. It takes 400 seconds to train the model and 0.0324 seconds for inference per test sample. We processed and converted the data in \textit{Care Record Data.zip} for training and \textit{Test Data.zip} for validation.

\subsection{Results and Discussion}

We trained the LSTM and BiLSTM models using our proposed model architecture in Figure \ref{arch}. First of all, different experiments were conducted for user-agnostic pre-training. Table \ref{pretrain_results} shows the performance of the conducted experiments.

The BiLSTM model pre-trained on the \nth{4} Nurse Care data shows the best performance in recall which identifies that the number of true positives among the actual positives is high. The LSTM model pre-trained on the same data, shows the best performance in precision and F1 score. 

The LSTM model pre-trained on both \nth{3} and \nth{4} Nurse Care data gives the best performance in accuracy. These pre-trained LSTM and BiLSTM models were used for further user-specific fine-tuning. 

\begin{table}[h]
\setlength{\tabcolsep}{6pt} 
\renewcommand{\arraystretch}{1.5}
\centering
\begin{tabular}{l|l|ccc|c}
\textbf{Model} & \textbf{Data} & \textbf{Accuracy} & \textbf{Precision} & \textbf{Recall} & \textbf{F1-score} \\ \hline
\textbf{BiLSTM} & 4th Nurse Care & 0.2414 & 0.4848 & \textbf{0.6781} & 0.5327 \\ \cline{2-2}
\textbf{LSTM} & 4th Nurse Care & 0.2452 & \textbf{0.5111} & 0.6776 & \textbf{0.5485} \\ \hline
\textbf{BiLSTM} & \begin{tabular}[c]{@{}l@{}}3rd and 4th \\ Nurse Care\end{tabular} & 0.2414 & 0.4848 & 0.6781 & 0.5327 \\ \cline{2-2}
\textbf{LSTM} & \begin{tabular}[c]{@{}l@{}}3rd and 4th \\ Nurse Care\end{tabular} & \textbf{0.2548} & 0.4908 & 0.6862 & 0.5453 \\ \hline
\end{tabular}
\caption{User-agnostic pre-training performance}
\label{pretrain_results}
\end{table}

\begin{table}[h]
\setlength{\tabcolsep}{6pt} 
\renewcommand{\arraystretch}{1.5}
\centering
\begin{tabular}{l|cc|ccc|c}
\textbf{Model} & \multicolumn{1}{l}{\textbf{user\_id}} & \textbf{\#valid\_samples} & \textbf{Accuracy} & \textbf{Precision} & \textbf{Recall} & \textbf{F1-score} \\ \hline
\textbf{LSTM} & \textbf{8} & 128 & 0.0703 & 0.5067 & 0.6353 & 0.5313 \\
 & \textbf{13} & 74 & 0.3108 & 0.5532 & 0.7005 & 0.5958 \\
 & \textbf{14} & 119 & 0.3333 & 0.6523 & 0.7582 & 0.6729 \\
 & \textbf{15} & 139 & 0.5286 & 0.6472 & 0.7674 & 0.675 \\
 & \textbf{25} & 66 & 0.3182 & 0.4758 & 0.5227 & 0.4786 \\ \cline{2-7} 
 & \textbf{avg} &  & \textbf{0.3158} & \textbf{0.5794} & \textbf{0.6931} & \textbf{0.6038} \\ \hline
\textbf{BiLSTM} & \textbf{8} & 128 & 0.0469 & 0.4459 & 0.6505 & 0.4985 \\
 & \textbf{13} & 74 & 0.3108 & 0.5491 & 0.7288 & 0.601 \\
 & \textbf{14} & 119 & 0.2583 & 0.6768 & 0.7019 & 0.6683 \\
 & \textbf{15} & 139 & 0.5143 & 0.6865 & 0.7618 & 0.6875 \\
 & \textbf{25} & 66 & 0.303 & 0.4192 & 0.596 & 0.4658 \\ \cline{2-7} 
 & \textbf{avg} &  & \textbf{0.2875} & \textbf{0.5729} & \textbf{0.6957} & \textbf{0.5972} \\ \hline
\end{tabular}
\caption{User-specific fine-tuning performance}
\label{finetuning_results}
\end{table}

During user-specific fine-tuning, we use a specific user's activity data for fine-tuning. Table \ref{finetuning_results}, shows the results of all the fine-tuning experiments. Different users had different sizes of validation samples. Hence, a weighted average is considered when aggregating the results.

Table \ref{finetuning_results} shows the lowest user-specific performance for user 8 even though it contains the highest number of training samples [Figure \ref{4th_dist}]. This indicates that the training and validation samples are inconsistent and noisy. Although user 15 contains the second lowest number of training samples, we observed the best user-specific performance for it. From these observations, it can be inferred that the overall dataset is very noisy because we can see the opposite of general deep learning model trends.

After aggregating the results as a weighted average, the table shows that LSTM fine-tuning achieved the best performance in accuracy, precision, and F1-score. These fine-tuned models are employed for final inference in the test dataset.

Comparing Table \ref{pretrain_results} and \ref{finetuning_results}, we can infer the following: 1) Adding \nth{3} Nurse Care Data with the \nth{4} Nurse Care data gave us better performance in fine-tuning. 2) User-agnostic fine-tuning would have resulted in worse performance than user-specific fine-tuning. 3) If we did not conduct the 2-stage training approach, our highest accuracy might have been around 24-25\%. Employing this approach, resulted in 5-6\% more accuracy and F1-score.

\section{Conclusion and Future Works}
The \nth{4} Nurse Care activity recognition challenge came up with a very unique and strenuous problem, but, if we consider the real-life implications, this task will facilitate increasing the efficiency of the nurse and caregivers and help them get rid of procrastination. This problem aims to predict the personalized probable activities of the next hour based on their activities in previous hours. We proposed an LSTM-based multi-label sequence prediction model and conduct a 2-stage training approach, i.e., user-agnostic pre-training and user-specific fine-tuning, to utilize both \nth{3} and \nth{4} nurse care activity recognition challenge datasets. Finally, we achieved the validation accuracy of 31.58\%, precision 57.94\%, recall 68.31\%, and F1 score 60.38\%.

Even though we achieved good performance with our 2-stage training approach, there are still some problems that we would like to pursue further in the future. First of all, the dataset was very imbalanced. We plan to work more on this problem to handle the data imbalance problem for care record data. Secondly, we would also like to explore how different data processing approaches would correlate with the performance of the task. Finally, we would also like to try some more advanced sequence models like transformers to compare the results with LSTM and BiLSTM.

\bibliographystyle{IEEEtran}
\bibliography{REF.bib}

\begin{thebibliography}{10}
\providecommand{\url}[1]{#1}
\csname url@samestyle\endcsname
\providecommand{\newblock}{\relax}
\providecommand{\bibinfo}[2]{#2}
\providecommand{\BIBentrySTDinterwordspacing}{\spaceskip=0pt\relax}
\providecommand{\BIBentryALTinterwordstretchfactor}{4}
\providecommand{\BIBentryALTinterwordspacing}{\spaceskip=\fontdimen2\font plus
\BIBentryALTinterwordstretchfactor\fontdimen3\font minus
  \fontdimen4\font\relax}
\providecommand{\BIBforeignlanguage}[2]{{%
\expandafter\ifx\csname l@#1\endcsname\relax
\typeout{** WARNING: IEEEtran.bst: No hyphenation pattern has been}%
\typeout{** loaded for the language `#1'. Using the pattern for}%
\typeout{** the default language instead.}%
\else
\language=\csname l@#1\endcsname
\fi
#2}}
\providecommand{\BIBdecl}{\relax}
\BIBdecl

\bibitem{inoue2019nurse}
S.~Inoue, P.~Lago, S.~Takeda, A.~Shamma, F.~Faiz, N.~Mairittha, and
  T.~Mairittha, ``Nurse care activity recognition challenge,'' \emph{IEEE
  Dataport}, vol.~1, p.~4, 2019.

\bibitem{vchd-s336-22}
\BIBentryALTinterwordspacing
S.~I. D. H. C. G. H. K. N. N. T. H. S. S. A.~P. Lago, ``4th nurse care activity
  recognition challenge datasets,'' 2022. [Online]. Available:
  \url{https://dx.doi.org/10.21227/vchd-s336}
\BIBentrySTDinterwordspacing

\bibitem{espinilla2018ucami}
M.~Espinilla, J.~Medina, and C.~Nugent, ``Ucami cup. analyzing the uja human
  activity recognition dataset of activities of daily living,''
  \emph{Multidisciplinary Digital Publishing Institute Proceedings}, vol.~2,
  no.~19, p. 1267, 2018.

\bibitem{lago2017contextact}
P.~Lago, F.~Lang, C.~Roncancio, C.~Jim{\'e}nez-Guar{\'\i}n, R.~Mateescu, and
  N.~Bonnefond, ``The contextact@ a4h real-life dataset of daily-living
  activities,'' in \emph{International and interdisciplinary conference on
  modeling and using context}.\hskip 1em plus 0.5em minus 0.4em\relax Springer,
  2017, pp. 175--188.

\bibitem{stikic2011weakly}
M.~Stikic, D.~Larlus, S.~Ebert, and B.~Schiele, ``Weakly supervised recognition
  of daily life activities with wearable sensors,'' \emph{IEEE transactions on
  pattern analysis and machine intelligence}, vol.~33, no.~12, pp. 2521--2537,
  2011.

\bibitem{10.1145/3460418.3479391}
\BIBentryALTinterwordspacing
S.~S. Alia, K.~Adachi, T.~Hossain, N.~T. Le, H.~Kaneko, P.~Lago, T.~Okita, and
  S.~Inoue, ``Summary of the third nurse care activity recognition challenge -
  can we do from the field data?'' in \emph{Adjunct Proceedings of the 2021 ACM
  International Joint Conference on Pervasive and Ubiquitous Computing and
  Proceedings of the 2021 ACM International Symposium on Wearable Computers},
  ser. UbiComp '21.\hskip 1em plus 0.5em minus 0.4em\relax New York, NY, USA:
  Association for Computing Machinery, 2021, p. 428–433. [Online]. Available:
  \url{https://doi.org/10.1145/3460418.3479391}
\BIBentrySTDinterwordspacing

\bibitem{10.1145/3341162.3344848}
\BIBentryALTinterwordspacing
M.~N. Haque, M.~Mahbub, M.~H. Tarek, L.~N. Lota, and A.~A. Ali, ``Nurse care
  activity recognition: A gru-based approach with attention mechanism,'' in
  \emph{Adjunct Proceedings of the 2019 ACM International Joint Conference on
  Pervasive and Ubiquitous Computing and Proceedings of the 2019 ACM
  International Symposium on Wearable Computers}, ser. UbiComp/ISWC '19
  Adjunct.\hskip 1em plus 0.5em minus 0.4em\relax New York, NY, USA:
  Association for Computing Machinery, 2019, p. 719–723. [Online]. Available:
  \url{https://doi.org/10.1145/3341162.3344848}
\BIBentrySTDinterwordspacing

\bibitem{10.1145/3410530.3414611}
\BIBentryALTinterwordspacing
S.~S. Alia, P.~Lago, K.~Adachi, T.~Hossain, H.~Goto, T.~Okita, and S.~Inoue,
  ``Summary of the 2nd nurse care activity recognition challenge using lab and
  field data,'' in \emph{Adjunct Proceedings of the 2020 ACM International
  Joint Conference on Pervasive and Ubiquitous Computing and Proceedings of the
  2020 ACM International Symposium on Wearable Computers}, ser. UbiComp-ISWC
  '20.\hskip 1em plus 0.5em minus 0.4em\relax New York, NY, USA: Association
  for Computing Machinery, 2020, p. 378–383. [Online]. Available:
  \url{https://doi.org/10.1145/3410530.3414611}
\BIBentrySTDinterwordspacing

\bibitem{10.1145/3410530.3414335}
\BIBentryALTinterwordspacing
M.~G. Rasul, M.~H. Khan, and L.~N. Lota, ``Nurse care activity recognition
  based on convolution neural network for accelerometer data,'' in
  \emph{Adjunct Proceedings of the 2020 ACM International Joint Conference on
  Pervasive and Ubiquitous Computing and Proceedings of the 2020 ACM
  International Symposium on Wearable Computers}, ser. UbiComp-ISWC '20.\hskip
  1em plus 0.5em minus 0.4em\relax New York, NY, USA: Association for Computing
  Machinery, 2020, p. 425–430. [Online]. Available:
  \url{https://doi.org/10.1145/3410530.3414335}
\BIBentrySTDinterwordspacing

\bibitem{irbaz2020nurse}
M.~S. Irbaz, A.~Azad, T.~A. Sathi, and L.~N. Lota, ``Nurse care activity
  recognition based on machine learning techniques using accelerometer data,''
  in \emph{Adjunct Proceedings of the 2020 ACM International Joint Conference
  on Pervasive and Ubiquitous Computing and Proceedings of the 2020 ACM
  International Symposium on Wearable Computers}, 2020, pp. 402--407.

\bibitem{10.1145/3351244}
\BIBentryALTinterwordspacing
S.~Inoue, P.~Lago, T.~Hossain, T.~Mairittha, and N.~Mairittha, ``Integrating
  activity recognition and nursing care records: The system, deployment, and a
  verification study,'' \emph{Proc. ACM Interact. Mob. Wearable Ubiquitous
  Technol.}, vol.~3, no.~3, sep 2019. [Online]. Available:
  \url{https://doi.org/10.1145/3351244}
\BIBentrySTDinterwordspacing

\bibitem{10.1145/3460418.3479388}
\BIBentryALTinterwordspacing
F.~R. Sayem, M.~M. Sheikh, and M.~A.~R. Ahad, ``Feature-based method for nurse
  care complex activity recognition from accelerometer sensor,'' in
  \emph{Adjunct Proceedings of the 2021 ACM International Joint Conference on
  Pervasive and Ubiquitous Computing and Proceedings of the 2021 ACM
  International Symposium on Wearable Computers}, ser. UbiComp '21.\hskip 1em
  plus 0.5em minus 0.4em\relax New York, NY, USA: Association for Computing
  Machinery, 2021, p. 446–451. [Online]. Available:
  \url{https://doi.org/10.1145/3460418.3479388}
\BIBentrySTDinterwordspacing

\bibitem{hj46-zs46-21}
\BIBentryALTinterwordspacing
S.~S. A. K. A. N. T. L. H. K. P. L.~S. Inoue, ``Third nurse care activity
  recognition challenge,'' 2021. [Online]. Available:
  \url{https://dx.doi.org/10.21227/hj46-zs46}
\BIBentrySTDinterwordspacing

\bibitem{inoue2019integrating}
S.~Inoue, P.~Lago, T.~Hossain, T.~Mairittha, and N.~Mairittha, ``Integrating
  activity recognition and nursing care records: the system, deployment, and a
  verification study,'' \emph{Proceedings of the ACM on Interactive, Mobile,
  Wearable and Ubiquitous Technologies}, vol.~3, no.~3, pp. 1--24, 2019.

\bibitem{hochreiter1997long}
S.~Hochreiter and J.~Schmidhuber, ``Long short-term memory,'' \emph{Neural
  computation}, vol.~9, no.~8, pp. 1735--1780, 1997.

\bibitem{schuster1997bidirectional}
M.~Schuster and K.~K. Paliwal, ``Bidirectional recurrent neural networks,''
  \emph{IEEE transactions on Signal Processing}, vol.~45, no.~11, pp.
  2673--2681, 1997.

\bibitem{kingma2014adam}
D.~P. Kingma and J.~Ba, ``Adam: A method for stochastic optimization,''
  \emph{arXiv preprint arXiv:1412.6980}, 2014.

\bibitem{loshchilov2017decoupled}
I.~Loshchilov and F.~Hutter, ``Decoupled weight decay regularization,''
  \emph{arXiv preprint arXiv:1711.05101}, 2017.

\end{thebibliography}
\end{document}